\documentclass{article}
\usepackage[utf8]{inputenc}

\usepackage{geometry}
 \newenvironment{acknowledgements}{%
  \section*{Acknowledgements}
}{%
}

\title{Gappy local conformal auto-encoders\\for heterogeneous data fusion: in praise of rigidity}

\usepackage{authblk}
\author[1,2]{Erez Peterfreund}
\author[3]{Iryna Burak}
\author[2]{Ofir Lindenbaum}
\author[4]{Jim Gimlett}
\author[3]{Felix Dietrich}
\author[2]{Ronald R. Coifman}
\author[1]{Ioannis G. Kevrekidis}
\affil[1]{Johns Hopkins University, Baltimore, MD, USA}
\affil[2]{Yale University, New Haven, CT, USA}
\affil[3]{Technical University of Munich, Munich, Germany}
\affil[4]{ISR Corp., Owings Mills, MD, USA}

\usepackage{amsthm}
\usepackage{xcolor}
\usepackage{ifthen}

\newcommand{\TODO}[1]{ 
\ifx\NOTES\undefined\else
{\color{red} [!]}\footnote{ {\color{red} TODO: #1}}
\fi
}

\newcommand{\NOTE}[1]{ 
\ifx\NOTES\undefined\else
  \footnote{ {\color{blue} NOTE: #1}}  
\fi
}

\newcommand{\ecomment}[1]{ 
\ifx\NOTES\undefined\else 
{\color{blue}[E]}\footnote{ {\color{blue} Erez: #1}}
\fi
}

\newcommand{\mcomment}[1]{ 
\ifx\NOTES\undefined\else
  {\color{green} [M]}\footnote{ {\color{green} Matan: #1}}  
\fi
}

\newcommand{\vast}{\bBigg@{4}}
\newcommand{\Vast}{\bBigg@{5}}

\newcommand {\V}[1] {{\mbox{\boldmath $#1$}}}

\newcommand{\Pd}[3]{\ifthenelse{\equal{#3}{1}}{\frac{\partial #1}{\partial #2}}{\frac{\partial^{#3} #1}{\partial #2^{#3}}}}

\usepackage{lineno}

\newcommand{\figref}[1]{Fig. \ref{#1}}
\usepackage{amssymb}
\usepackage{hyperref} 

\usepackage{algorithm2e}

\usepackage{caption}
\usepackage{subcaption}

\usepackage{float}
\usepackage{graphicx}
\usepackage{amsmath}
\usepackage{wrapfig}
\usepackage{verbatim}
\usepackage{hhline}
\usepackage{cancel}
\usepackage{mathabx}

\usepackage{xcite}

\graphicspath{{../pdf/}{../jpeg/}{/graphs/}}    
\DeclareGraphicsExtensions{.pdf,.jpeg,.png,.jpg}

\newcommand{\latentdomain}{\mathcal{X}}
\newcommand{\latentdomaink}[1]{\mathcal{X}^{#1}}

\newcommand{\measureddomaink}[1]{\mathcal{Y}^{#1}}

\newcommand{\rvlatent}[2]{\V{X}^{#1}_{#2}}
\newcommand{\rvmeasured}[2]{\V{Y}^{#1}_{#2}}
\newcommand{\measuredfunctionk}[1]{\V{f}^{#1}}

\newcommand{\burstcenterlatent}[2]{\V{x}^{#1}_{#2}}
\newcommand{\burstcentermeasured}[2]{\V{y}^{#1}_{#2}}
\newcommand{\burstsample}[3]{\V{y}^{#1}_{#2, #3}}

\theoremstyle{definition}

\begin{document}

\maketitle

\begin{abstract}

       Fusing measurements from multiple, heterogeneous, partial sources, observing a common object or process,  poses challenges due to the increasing availability of numbers and types of sensors.
       In this work we propose, implement and validate an end-to-end computational pipeline in the form of a multiple-auto-encoder neural network architecture for this task.
The inputs to the pipeline are several sets of partial observations, and the result is a globally consistent latent space, harmonizing (rigidifying, fusing) all measurements.
The key enabler is the availability of multiple slightly perturbed measurements of  each instance:, local measurement, ``bursts", that allows us to estimate the local distortion induced by each instrument.
We demonstrate the approach in a sequence of examples, starting with
simple two-dimensional data sets and proceeding to a Wi-Fi localization problem and to the solution of a ``dynamical puzzle" arising in spatio-temporal observations of the solutions of Partial
Differential Equations.

\end{abstract}

\section{Introduction}
In recent years, the availability of high-dimensional data from diverse sources, including novel types of sensors, has amplified the need for reliable data integration methods. In scientific experiments, for example, we often record the behavior of a physical system using different measurement instruments. Each instrument successfully resolves a different subset of the system observables over a certain scale. The integration of all the observations should provide us with a better understanding of the system and its behavior; 
the need for such integration arises in many fields of science, from biology and chemistry to computer science (e.g., computer vision and natural language processing).
When decision models are based on multiple types of observations as input (such as RADAR, LIDAR, camera, and ultrasonic sensors in autonomous vehicles), \textit{sensor fusion} becomes critical.

Specific approaches to data integration use methods from \textit{manifold learning} and \textit{dimensionality reduction}. Techniques from these fields could lead to an informative reparameterization of observations from different (groups of) sensors; when each sensor / group of sensors only observes a part of the entire system behavior, systematically integrating multiple representations into one coherent picture remains challenging. In machine learning, there is considerable interest in methods that can accomplish this, e.g., in text-to-image models \cite{chen2020simple, radford2021learning}.

\begin{figure}[ht!]
    \centering
    \includegraphics[width=.6\textwidth]{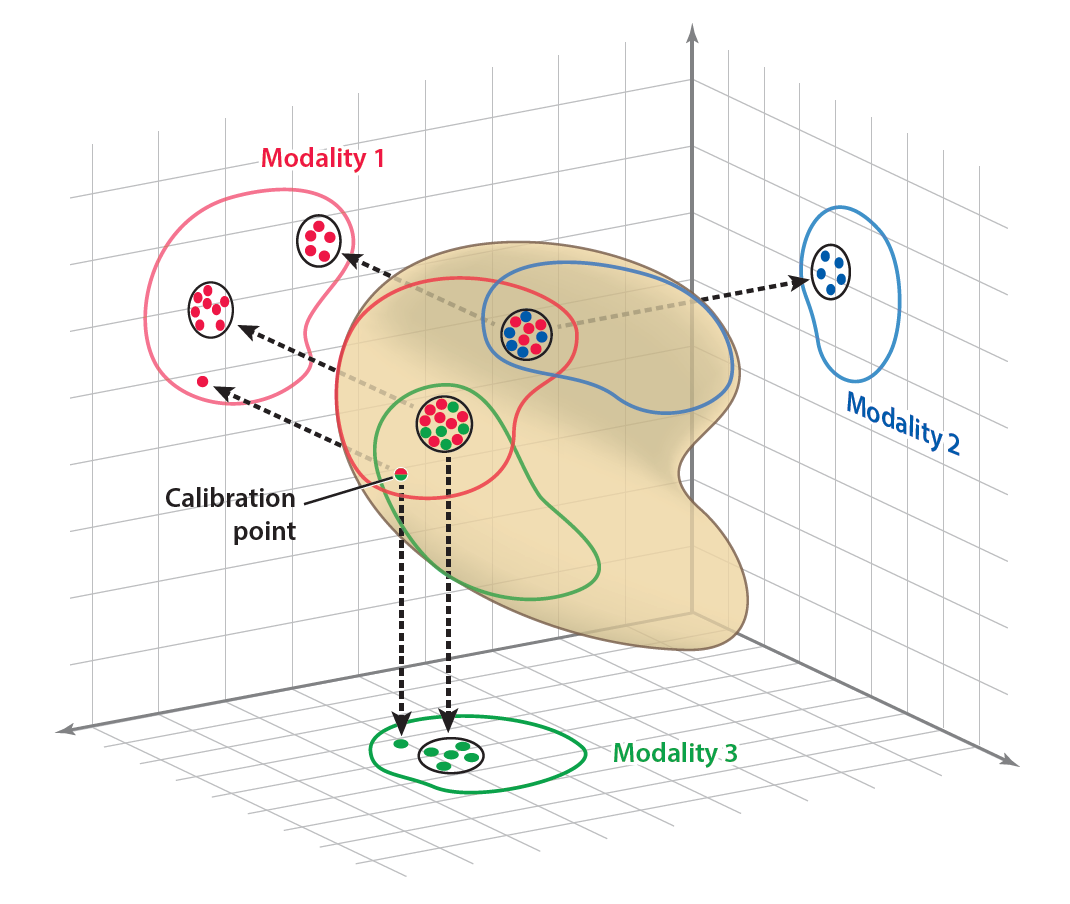}
    \caption{The latent representation of the data (center, three-dimensional) is observed by three different modalities on their respective two-dimensional coordinate planes. Each modality captures only a subset of the latent domain and has its own unique deformation. Local point neighborhoods of points in the latent space are transformed into ellipses in the modality spaces. A few points in the intersection regions are being observed by more than one modality. These are the \textit{calibration points} used to rigidly combine (and thus fuse) the different modalities. }
    \label{fig:lat_modalities}
\end{figure}

This paper presents a deep learning based method towards systematic sensor integration, for the case where sensors only partially observe the system.
It extends previous work~\cite{peterfreund2020local}, which fused complete (as opposed to partial) sensor information.
We will study data that are assumed to lie on a path-connected manifold in an \textit{inaccessible, latent} state space (illustrated in \figref{fig:lat_modalities}). This data is observed through different measurement devices, each of them modeled as a smooth unknown non-linear function of the system states. Additionally, to enable registration of different sets of partial observations, we require a small set of ``common points'', i.e., overlapping system states, measured by at least two different devices. 
The objective is to find a parameterization for the measured data that creates a useful realization of the global latent manifold. 

The key enabling feature of the approach, both in previous work and here, is the (assumed) ability to observe system states using a perturbed sampling strategy as suggested in \cite{singer2008non} and used in other studies \cite{singer2009detecting,talmon2013empirical, wu2014assess,dsilva2016data,kemeth2018emergent,peterfreund2020local}. The sampling strategy provides the scientist not just with isolated samples, but also a set of observed ``bursts'' around each sample. A ``burst" is a set of states located in the neighborhood of each sample point. This neighborhood is observed by each of the locally available sensors. Our method hinges on exploiting this additional information, and allows us to build a consistent parameterization of the inaccessible latent manifold. 

Extending this approach to a setting with multiple sensors, each with partial information, is challenging. The seminal work of~\cite{singer2008non} on ``anisotropic Diffusion Maps'', assumes access to all sampled data points over the entire latent manifold (see \figref{fig:example_missing_dists}).
In this work, the sensors only have partial information. We will first consider what we call the ``patch case'':
Here, sets of sensors jointly observe entire connected parts of the latent manifold. The difficulty then lies in registering these patches and combining them into one rigid representation; in particular, in determining and exploiting the \textit{minimal} number of common points across the patches.
We will also mention (and discuss in Appendix~\ref{appendix:point case}), what we call the ``point case''. Here, each sensor only observes single points (and the burst around them). Now, the difficulty lies in determining the minimal number of ``common'' sensors that simultaneously observe each point, in an effort to combine all \textit{points} (as opposed to all patches) into one rigid representation.
%
%
%

For the ``patch case'', we propose a neural network based solution composed of a set of encoder-decoder pairs for each patch. Each encoder maps its patch into a latent space that is common for all encoders. The corresponding decoder ensures that no information is lost. Moreover, in order to register the embeddings of the various encoders, we exploit our  set of common points.
A similar approach will be used for the ``point case''.

The remainder of the paper is organized as follows.
In section~\ref{sec:mathematical framework} we introduce in detail our proposed network architecture, employing multiple simultaneous auto-encoders to integrate partial information;
its properties are analyzed in section~\ref{sec:analysis}.
In section~\ref{sec:numerical experiments}, we demonstrate the capability of registering multiple partial sensor measurements in two applications (1) a Wi-Fi sensor fusion problem, and (2) the integration of multiple partial observations of the solution of a partial differential equation over a complicated, two-dimensional domain (inspired by voting districts in a US state).
We conclude in section~\ref{sec:discussion}, by discussing the limitations of the approach as well as potential future work.

\section{Mathematical framework}\label{sec:mathematical framework}

 We will start with an intuitive explanation of the problem (illustrated in \figref{fig:lat_modalities}). Consider a large set of entities (here, points), each of which is observed (measured) by a certain number of \textit{scalar-valued} sensors: the axes in the figure.
 We assume that the points lie on a $d$-dimensional smooth manifold in the $K$-dimensional space of sensors. We know from Whitney's theorem that if all points are observed by a subset of $2d+1$ (generic) \textit{common} sensors, the sensor measurements provide an embedding of this manifold.
 We will refer to such a subset of sensors rich enough to embed the data as a ``modality''.
 We are interested in the case where not all points are simultaneously observed by the same modality, and we seek the minimum conditions under which heterogeneous sensor  observations can be fused.
 In previous work, we showed how to create an embedding invariant (modulo isometries) to changing the modality. This naturally brings up the question of fusing these modalities, creating a common embedding for all of them, and that is the purpose of this paper.
 The main tool that enabled invariance to changing the modality was the observation not just of each point, but of a ``burst'' around it, meaning observations of a small neighborhood of the point on the manifold. This was motivated by the original, anisotropic Diffusion Maps work of Singer and Coifman~\cite{singer2008non}.


We assume the latent space we seek is a $d$-dimensional path connected manifold $\mathcal{X}$. We also assume that $K$ different modalities capture specific regions of $\mathcal{X}$, denoted by $\latentdomaink{1},\ldots,\latentdomaink{K}\subset \latentdomain$, and that the union of these regions is path-connected. 
These modalities are described by the non-linear functions $\measuredfunctionk{1}:\latentdomaink{1}\rightarrow \measureddomaink{1},\ldots,\measuredfunctionk{K}:\latentdomaink{K}\rightarrow \measureddomaink{K}$,
where $\measureddomaink{1}\subset\mathbb{R}^{D_1},\ldots,\measureddomaink{K}\subset\mathbb{R}^{D_K}$ are defined to be the observations/modalities spaces and $D_1,\ldots,D_k\geq 2d+1$ (it may be that as low as $d$ would suffice, but $2d+1$ is necessary to guarantee embeddings). Furthermore, we assume  $\measuredfunctionk{1},\ldots,\measuredfunctionk{K}$ to be smooth and injective. Our goal is to find an embedding in $\mathbb{R}^{2d+1}$ (or, if we are lucky, as low as $\mathbb{R}^{d}$) that is isometric to $\bigcup_i \latentdomaink{i}$ from samples captured in $\measureddomaink{1},\ldots,\measureddomaink{K}$ .

Each modality with index $k\in \{1,\ldots,K\}$ captures $N_k$ data points, denoted by $\burstcenterlatent{k}{1},\ldots, \burstcenterlatent{k}{N_k}\subset \latentdomaink{k}$. These data points are pushed forward to their respective image $\burstcentermeasured{k}{i}=\measuredfunctionk{k}(\burstcenterlatent{k}{i})$ for $i=1,\ldots,N_k$. We assume that instead of only observing $\burstcenterlatent{k}{i}$ (and obtaining $\burstcentermeasured{k}{i}$), a burst in the form of $M$ samples around $\burstcenterlatent{k}{i}$ is captured. Formally, let the observations $\burstsample{k}{i}{1},\ldots \burstsample{k}{i}{M}$ be independent and identically distributed samples of the random variable
\begin{eqnarray*}
\rvmeasured{k}{i} = \measuredfunctionk{k} (\rvlatent{k}{i}),
\end{eqnarray*}
where $\rvlatent{k}{i}\sim \mathcal{N}_d(\burstcenterlatent{k}{i}, \sigma^2I_d)$ for $i=1,\ldots,N_k$ and $k=1,\ldots,K$. It is important to note that $\V{X}_i^k$ can attain other light-tailed isotropic distributions, such as a uniform distribution, over a ball around $\V{x}_i^k$ with an $\epsilon$ radius, $B_{x_i}(\epsilon)$. Hence, for each modality $k\in \{1,\ldots,K\}$ a set of $N_k$ bursts is captured in the form of $\{\burstsample{k}{i}{j}\}_{j=1,\ldots,M}$, for $i=1,\ldots,N_k$. Throughout the paper we will abuse notation and define $\rvmeasured{k}{i}=\{\burstsample{k}{i}{j}\}_{j=1,\ldots,M}$, unless indicated otherwise. 

\begin{figure}[H]
    \centering
    
    \includegraphics[width=.9\textwidth]{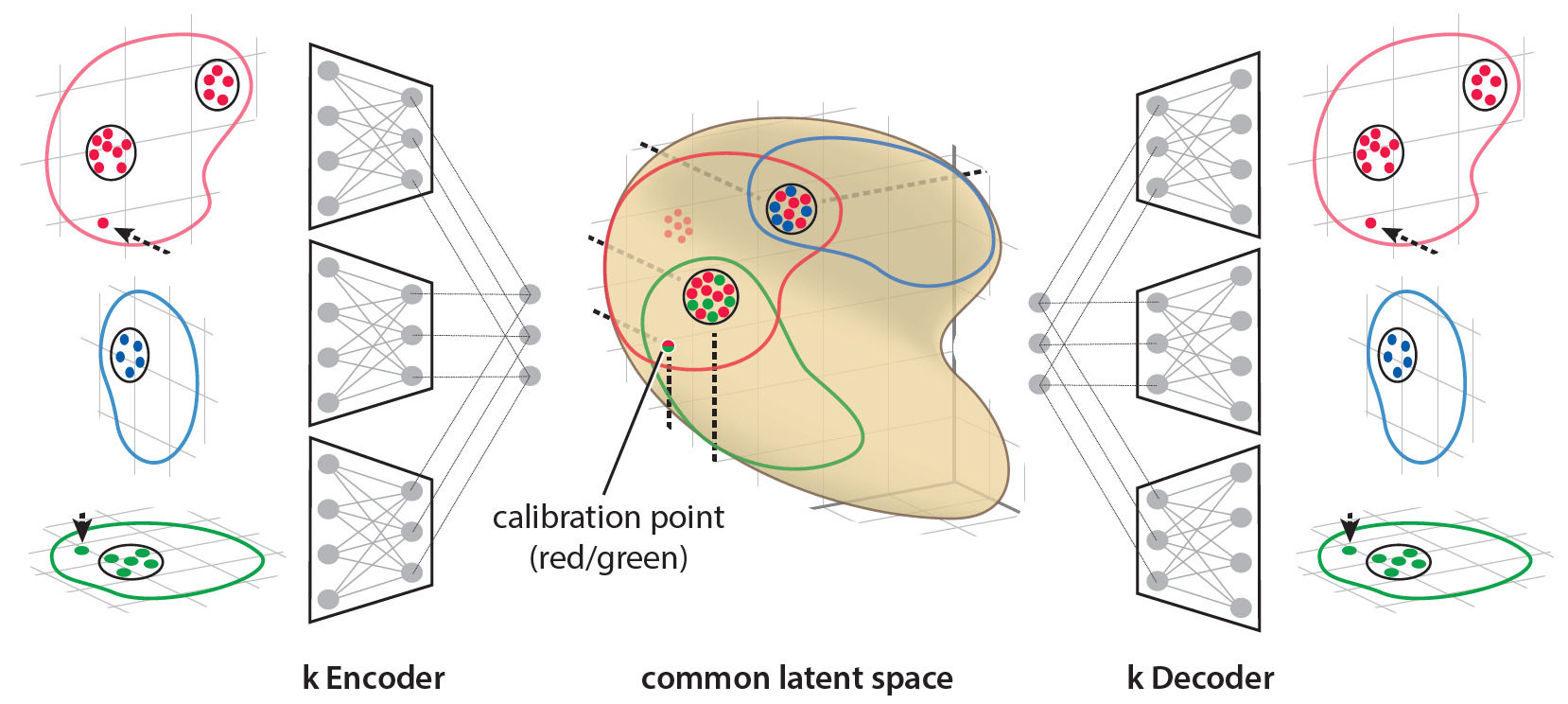}
    \caption{Gappy LOCA model trained on all the modalities simultaneously to form one shared embedding space. The intersecting regions are trained to match in the shared embedding space. The calibration points are necessary to fix any unresolved rigid body transformations between the individual embeddings.
    }
    \label{fig:loca architecture}
\end{figure}

Furthermore, to relate the different modalities across themselves, we assume to know of ``a few'' calibration bursts, i.e., bursts captured by more than one modality. More specifically, calibration bursts are pairs of bursts captured by two different modalities, where each pair shares the same burst center in the latent domain. We denote the indices of observations in calibration bursts in the form of quadruplets $(i,j,k,s)$ derived from $Y^k_i$ and $Y^s_j$.
In what follows, we will formulate a neural network architecture that will allow us to consistently fuse the measurements of different modalities in a common latent space in which all partially observed points are embedded. 
Fig.~\ref{fig:loca architecture} helps visualize our approach: an encoder and a decoder for each modality are connected through a shared latent space, in which the different observations are rigidly combined, exploiting the common registration points/bursts.
We will discuss below the minimal number of registration points required for such a rigid assembly.
We name this process ``Gappy LOCA''; the LOCA comes from our previous work on locally conformal auto-encoders, allowing isometric embeddings for each modality separately.
The word ``Gappy'' for the data is used by analogy to the literature on ``Gappy PODs''~\cite{carlberg-2010} and, recently, ``Gappy Diffusion Maps''~\cite{martin-linares-2023}.

The Gappy LOCA algorithm is based on $K$ auto-encoders, $\{( \V{\rho}^k, \V{\gamma}^k)\}$, with a shared embedding space.  Each encoder maps the modality's data into this shared embedding space, $\V{\rho}^k:\mathcal{Y}^k\rightarrow \mathbb{R}^d$, and each decoder maps the data back to the modality's space, $\V{\gamma}^k:\mathbb{R}^d\rightarrow \mathcal{Y}^k$. The training procedure of Gappy LOCA is built upon minimizing three loss terms: \textit{Whithening loss}, \textit{Reconstruction loss}, and \textit{Calibration loss}; the first two of these arose in \cite{peterfreund2020local}. The \textit{Whitening loss} aims to find a common, coherent embedding for the different modalities and exploits the information contained in the local measurement bursts. It is defined by 
\begin{eqnarray*}
L_{white}(\{ \V{\rho}^k\}) &=& \sum_{k=1}^k L_{white}^k( \V{\rho}^k)\\
L_{white}^k ( \V{\rho}^k) &=& \left \| \frac{1}{\sigma^2}\hat{\V{C}} \left( \V{\rho}^k( \V{Y}_i^k) \right) -\V{I}_d \right\|_F^2,
\end{eqnarray*}
where $\V{Y}_i^k$ is the $i-$th burst of the $k-$th observed modality, and $\hat{\V{C}} \left( \V{\rho} ^k( \V{Y}_i^k) \right)$ is the empirical covariance of this burst after each of its samples were pushed forward through $\V{\rho}^k$. It is important to note that in \cite{singer2008non} a direct connection was established between the Jacobian $J_{f^k}(x_i^k)$ and the covariance of $\V{Y}_i^K$. We use this connection to establish a relation between the original latent space $\mathcal{X}$ and its embedding function here understood as $\V{\rho}^k\circ \V{f}^k$.

While the $K$ encoders provide a shared embedding space for all the different modalities, each decoder allows us to go back to the individual modality space, ensuring that no information has been lost.
We define the reconstruction function by
\begin{eqnarray}
L_{recon} &=& \sum_{k=1}^K L_{recon}^k( \V{\rho}^k, \V{\gamma}^k)\\
L^k_{recon} (\V{\rho}^k, \V{\gamma}^k) &=&  \frac{1}{N_k} \sum_{i=1}^{N_k} \sum_{\V{y}\in \V{Y}_i^k} \frac{\left\| \V{y} - \V{\gamma}^k( \V{\rho}^k(\V{y}) ) \right\|_2^2}{\lambda_k \cdot D_k},
\end{eqnarray}
where $\lambda_k$ equals to the median over the $d-$th eigenvalue of $\hat{C}(Y^k_1),\ldots, \hat{C}(Y^k_{N_k})$. The fact that  $\lambda_k$ is modality dependent allows relating the different losses to each other. 

Introducing the \textit{Calibration loss} becomes necessary for registering the several, partial observations of the data in a consistent, rigid way; We remind the reader that there are important conditions on the minimal number of calibration points that would allow this to be performed successfully.
We define the \textit{Calibration loss} by
\begin{eqnarray}
L_{calib} &=&   \frac{1}{|I|} \sum_{(i,j,k,s)\in I}  \frac{ \| \V{\rho}^k _i - \V{\rho}^s_j\|_2^2}{ d\cdot \sigma^2}, 
\end{eqnarray}
where $\V{\rho}^s_i = (1/m) \sum_{\V{y}\in \V{Y}^s_i} \V{\rho}^s(\V{y})$ is the approximated burst's center embedding.

\section{Analysis}\label{sec:analysis}

We now provide a brief analysis of the data requirements of Gappy LOCA, and provide an algorithm to determine if the requirements are satisfied. In the analysis, we distinguish the ``patch case" and the ``point case".

\subsection{Patches connected by common points}\label{sec:patches-common}
The objective is to combine $N$ rigid bodies $B_i\subset\mathbb{R}^D$, $i=1,\dots,N$, in a $D>1$ dimensional Euclidean space.
We assume all bodies $B_i$ are path-connected manifolds, all of which have the same intrinsic dimension $d\leq D$.
Also, we assume access to a (sufficient, to be discussed) set of pairs of ``connection points'' $(x_l,x_k)$, $k\neq l$,
where $x_l\in B_l$ is a point inside body $B_l$ and $x_k\in B_k$ is a point inside another body $B_k$, that we know should coincide in the combined state, i.e. $x_l=x_k$ for all such pairs.
We consider each pair $(x_l,x_k)$ as an edge $e_i\in V$, and each body $B_l$ as a vertex $v_l\in V$, in a graph $G=(V,E)$.
The $N$ rigid bodies can be ``rigidified'', i.e., they can be combined uniquely---up to isometry of the combination in the ambient space---into a single path-connected manifold $B=\cup_{i=1}^N B_i$, if the following requirements are satisfied:
\begin{enumerate}
    \item It must be possible to reach all bodies from any starting body by moving through the connections. In other words, the graph $G$ must not contain disconnected subgraphs (as checked by Alg.~\ref{alg:check_pairs}).
    \item The number of connection points \emph{per body} $B_i$ must be at least $d(d+1)/2$ to fix the degrees of freedom of orthogonal transformations (here: rigid body motion, i.e., rotations and reflections). Note that this only depends on the intrinsic dimension of the bodies, not the ambient space dimension.
\end{enumerate}
Checking requirement (1) is possible with a depth-first-search on a graph in $O(V+E)$ time,  where $V$ is the number of vertices (here: rigid bodies) and $E$ is the number of edges (here: connection pairs); cf.~Alg.~\ref{alg:check_pairs}.

\begin{algorithm}[H]
  \SetKwInOut{Input}{Input}
  \SetKwInOut{Output}{Output}
  
  \Input{graph}
  \Output{disconnectedSubgraphs}
  \caption{\label{alg:check_pairs}Detect disconnected subgraphs using depth-first-search.}
  
  \SetKwFunction{DepthFirstSearch}{DepthFirstSearch}
  \SetKwProg{Fn}{Function}{:}{}
  
  \Fn{\DepthFirstSearch{vertex, visited, disconnectedSubgraph}}{
    visited $\leftarrow$ visited $\cup$ \{vertex\}\;
    disconnectedSubgraph $\leftarrow$ disconnectedSubgraph $\cup$ \{vertex\}\;
    
    \For{neighbor in vertex.neighbors}{
      \If{neighbor $\notin$ visited}{
        \DepthFirstSearch{neighbor, visited, disconnectedSubgraph}\;
      }
    }
  }
  
  disconnectedSubgraphs $\leftarrow$ []\;
  visited $\leftarrow$ $\emptyset$\;
  
  \For{vertex in graph.vertices}{
    \If{vertex $\notin$ visited}{
      disconnectedSubgraph $\leftarrow$ $\emptyset$\;
      \DepthFirstSearch{vertex, visited, disconnectedSubgraph}\;
      disconnectedSubgraphs $\leftarrow$ disconnectedSubgraphs $\cup$ \{disconnectedSubgraph\}\;
    }
  }
  
  \Return{disconnectedSubgraphs}\;
\end{algorithm}

\subsection{Points connected by common sensors}\label{sec:points}
In a slightly different setting, the objective is to combine $N$ distinct points $p_i$, each measured by a list of associated scalar sensors $S_i=(s_{i,1},s_{i,2},\dots)$. We assume the sensors  are real-valued, smooth functions of a common manifold of intrinsic dimension $d$. For any pair of points $(p_l, p_k)$, we assume that it is possible to check if some the elements in the associated sensor sets $S_l,S_k$ are ``common'', s.t. $s_{l,i}=s_{k,j}$ for some $i,j$.
We can construct a graph analogous to the one in Sec.~\ref{sec:patches-common}, if we now define \textit{the points} as its vertices and \textit{the common sensors} as its edges: Two vertices are connected through an edge if they share at least $2d+1$ common sensors. This number of sensors is sufficient to establish a (Mahalanobis) distance between the two corresponding points in the latent space we will eventually construct.
Alg.~\ref{alg:check_pairs} can be used again to determine whether all points lie in a common subgraph.
The orthogonality requirement can also be stated in this context. If every point has at least $2d+1$ different sensors associated to it, there also must be at least $d(d+1)/2$ neighbors to fix the orthogonality constraints.

\section{Numerical experiments}\label{sec:numerical experiments}

Throughout this section we examine the quality of joint embeddings produced by the proposed algorithm in various settings. We compare our method to previous work~\cite{peterfreund2020local,singer2008non} that were adapted here to the multi-modality setting (see section~\ref{sec:mathematical framework}).
We start by demonstrating the Gappy LOCA architecture in cases where the latent space is simply $\mathcal{X}=\mathbb{R}^2$, and is observed by two modalities $\V{f}^{1},\V{f}^{2}:\mathcal{X}\rightarrow \mathbb{R}^2$, defined by
\begin{eqnarray*}
\V{f}^{1}(\V{x}) &=& \left( \begin{array}{c} 
x[1]\\
cos(2 \pi \cdot 0.3\cdot x[1])+x[2]
\end{array}\right), \qquad 
\V{f}^{2}(\V{x}) = \left( \begin{array}{c} 
x[1]\\
cos(\pi/2 + 2 \pi \cdot 0.3\cdot x[1])+x[2]
\end{array}\right).
\end{eqnarray*}
The domain of $\V{f}^{1},\V{f}^{2}$ as well as the amount of points (burst centers) varies throughout the different experiments, but the burst model persists across all examples: Given a point (burst center) $\V{x}_i^k$ in the latent space, for $k\in\left\lbrace 1,2\right\rbrace$ and $i=1,\ldots,N_k$, we construct a burst around it by using $m=100$ nearby points sampled independently from $\mathcal{N}(\V{x}_i^k,\sigma^2 \V{I}_2)$, where $\sigma=0.1$. 
Observations of all $m$ points through $\V{f}^{1},\V{f}^{2}$ provide the training data to the Gappy LOCA network.

The performance measure that we will use throughout the numerical experiments determines to what extend our constructed embedding differs from an isometric embedding of the original latent domain. Specifically, we compare Euclidean distances between any pair of points in the original latent domain to those in the reconstructed, joint embedding domain. For the LOCA-based embedding algorithms, we keep the embedding space as is, while we rescale the embedding of A-dmaps-based embeddings. 
We rescale the A-dmap embeddings to match the original scales.

\subsection{Experiment 1: One common domain}
We start by evaluating the different methods in a simplified setting, in which all modalities observe points in \textit{the same region} (domain) in the latent space. Specifically, we define the latent domains as $\mathcal{X}^1=\mathcal{X}^2= [0,\pi]^2$, from which we sample $500$ uniformly for each modality ($N_1=N_2=500$). Additionally, we sample three pairs of registration bursts, independently for each modality.

\begin{figure}[H]
    \centering
          \begin{subfigure}[b]{0.245\textwidth}
         \centering
         \includegraphics[width=\textwidth]{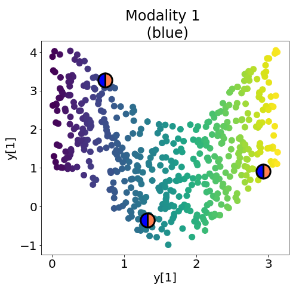}
         \caption{}
     \end{subfigure}
          \begin{subfigure}[b]{0.245\textwidth}
         \centering
         \includegraphics[width=\textwidth]{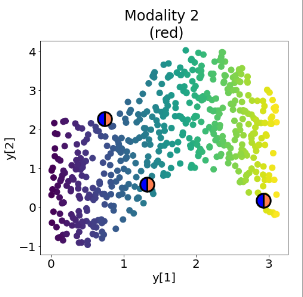}
         \caption{}
     \end{subfigure}
          \begin{subfigure}[b]{0.245\textwidth}
         \centering
         \includegraphics[width=\textwidth]{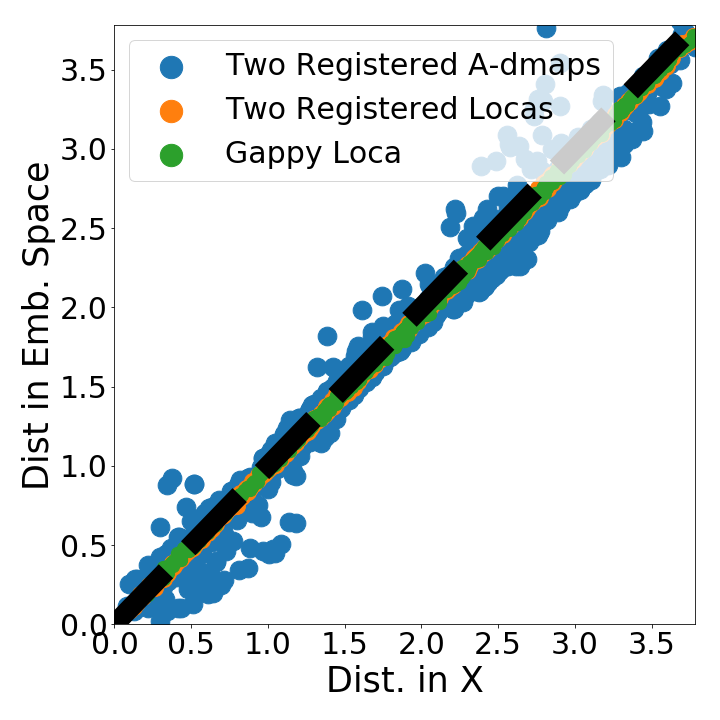}
         \caption{}
     \end{subfigure}
     \begin{subfigure}[b]{0.245\textwidth}
         \centering
         \includegraphics[width=\textwidth]{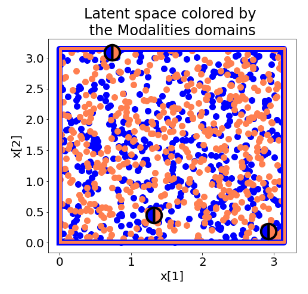}
         \caption{}
     \end{subfigure}
\caption{
(a)  Modality $1$ colored by $x[1]$. (b) Modality $2$ colored by $x[1]$. (c) Isometry performance graph, including a comparison between Euclidean distances across pairs of points in the latent space and in the embedding spaces of the different algorithms. (d) Points sampled in the latent domain colored by their modality. The red and blue frames indicate each modality's latent domain, which is the same in this experiment. }
\label{fig:experiments_same_domain}
\end{figure}

As demonstrated in Figure \ref{fig:experiments_same_domain}, Gappy LOCA, as well as two independent applications of LOCA to each modality followed by common point registration, both  provide an isometric embedding of the latent space. The embedding provided by A-dmaps is slightly less consistent (farther from isometric) to the original square in the latent space. A-dmaps suffers from deformations close to the boundary; registration bursts located close to the boundary distort the embedding.

Before we continue with our examples, we want to note something that was observed in training our networks. It can easily happen, depending on the initialization, that as the training progresses, an auto-encoder builds up one of the possible flips of its patch; it then becomes difficult for the registration component of the loss to coherently ``flip'' the entire patch-representation back, causing the training to stall.
We will return to this training artifact and suggest a possible remedy in the discussion.

\subsection{Experiment 2: Two domains with an overlap}
Here we evaluate the different algorithms on modalities that were sampled from two different domains, with an overlapping region between the two. We define the domains by $\mathcal{X}^1=[0,\pi]^2$ and $\mathcal{X}^2=[\pi/2,3\pi/2]\times [0,\pi]$, and sample $500$ points uniformly using each modality. Furthermore, we sample three points from the overlapping region for registration.

\begin{figure}[H]
    \centering

  \begin{subfigure}[b]{0.245\textwidth}
 \centering
 \includegraphics[width=\textwidth]{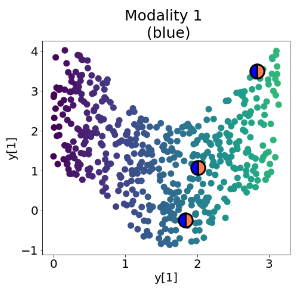}
 \caption{}
\end{subfigure}
  \begin{subfigure}[b]{0.245\textwidth}
 \centering
 \includegraphics[width=\textwidth]{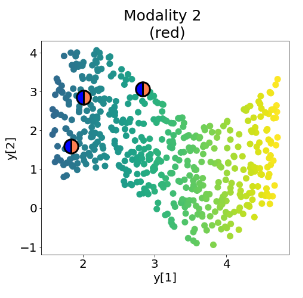}
 \caption{}
\end{subfigure}
  \begin{subfigure}[b]{0.245\textwidth}
 \centering
 \includegraphics[width=\textwidth]{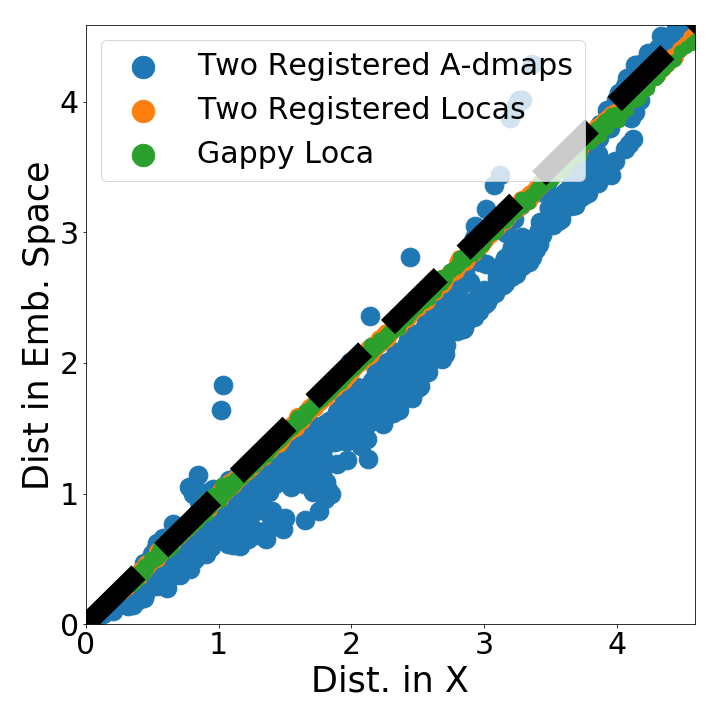}
 \caption{}
\end{subfigure}
\begin{subfigure}[b]{0.245\textwidth}
 \centering
 \includegraphics[width=\textwidth]{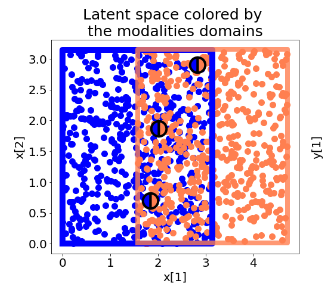}
 \caption{}
\end{subfigure}
\caption{(a)  Modality $1$ colored by $x[1]$. (b) Modality $2$ colored by $x[1]$. (c) Isometry performance graph, including a comparison between Euclidean distances across pairs of points in the latent space and in the embedding spaces of the different algorithms. (d) Points sampled in the latent domain colored by their modality. The red and blue frames indicate each modality's latent domain, which is slightly displaced in this experiment.}
\label{fig:experiments_nonsame_domain}
\end{figure}

In Figure \ref{fig:experiments_nonsame_domain} we show the configuration of the data along with a comparison of the method's capabilities to embed isometrically. Gappy LOCA and (repeated, and then registered) LOCA both lead to a near isometric embedding; the embedding provided by A-dmaps is again slightly less isometric in its latent space. 

\subsection{Experiment 3: A single modality with a patchy domain}
In this experiment, we work with modalities that produce one connected set and one disconnected set of observations. Disconnected sets raise issues for the different algorithms, as the algorithms attempt to connect the disjoint parts through different types of out-of-sample extensions between these regions. We define $\mathcal{X}^1=[0,\pi]^2\cup [2\pi,3\pi]\times [0,\pi]$ and $\mathcal{X}^2=[\pi/2,5\pi/2]\times [0,\pi]$, so that the intersection will be defined by $[\pi/2,\pi]\times [0,\pi]\cup [2\pi,5\pi/2]\times [0,\pi]$. We sample $1000$ points from $\mathcal{X}_1$ ($500$ out of each path-connected component) and $500$ from $\mathcal{X}_2$ using a uniform distribution over their domains. Furthermore, we sample six additional points, three from each (path-connected) intersection region.

\begin{figure}[H]
    \centering
  \begin{subfigure}[b]{0.245\textwidth}
 \centering
 \includegraphics[width=\textwidth]{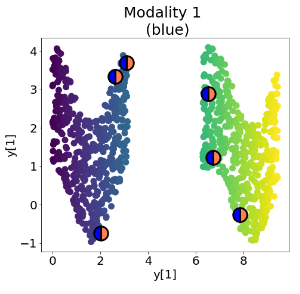}
 \caption{}
\end{subfigure}
  \begin{subfigure}[b]{0.245\textwidth}
 \centering
 \includegraphics[width=\textwidth]{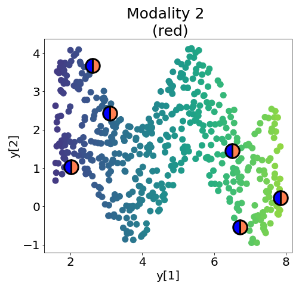}
 \caption{}
\end{subfigure}
  \begin{subfigure}[b]{0.245\textwidth}
 \centering
 \includegraphics[width=\textwidth]{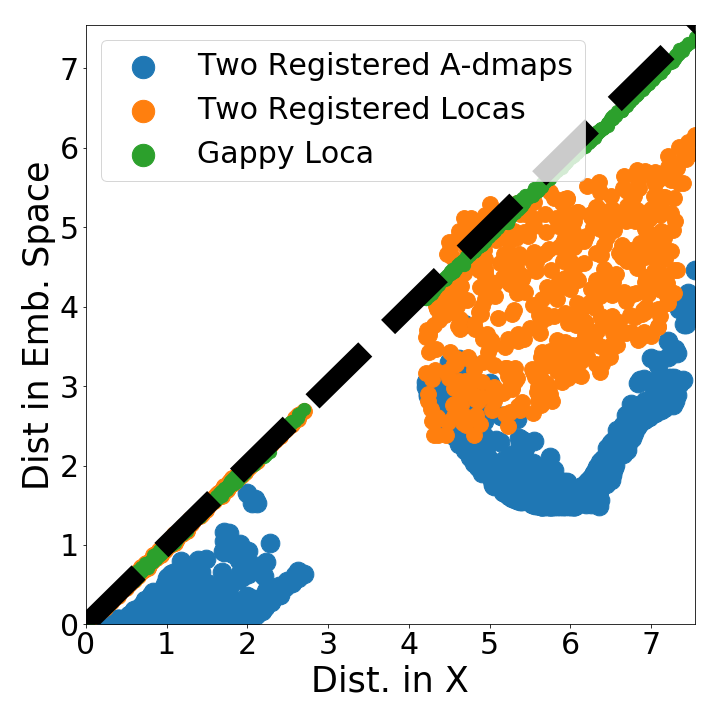}
 \caption{}
\end{subfigure}
  \begin{subfigure}[b]{0.245\textwidth}
 \centering
 \includegraphics[width=\textwidth]{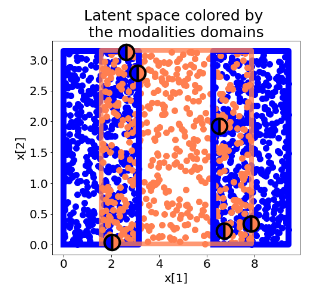}
 \caption{}
\end{subfigure}
\caption{(a) The two parts of the domain of Modality $1$ colored by $x[1]$. (b) Modality $2$ colored by $x[1]$. (c) Isometry performance graph, including a comparison between Euclidean distances across pairs of points in the latent space and in the embedding spaces of the different algorithms. Only Gappy LOCA successfully recovers the isometry. (d) Points sampled in the latent domain colored by their modality. The red and blue frames indicate each modality's latent domain, which is disconnected for one of the modalities in this experiment.}
\label{fig:experiments_patchy_domain}
\end{figure}

\begin{figure}[H]
    \centering
    \begin{subfigure}[t]{.49\linewidth}
    \includegraphics[width=1.\textwidth]{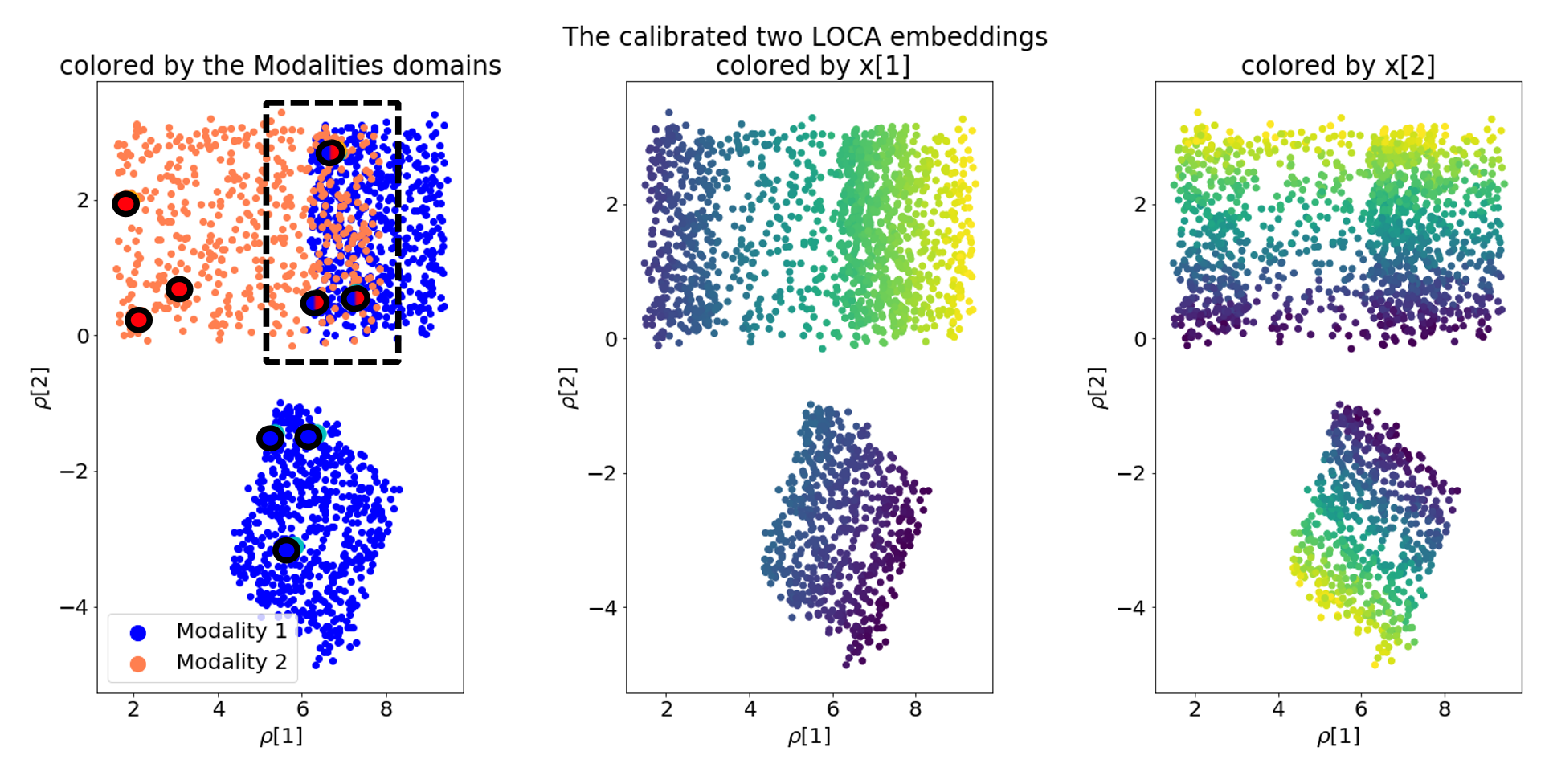}
    \caption{Two registered LOCAs embedding}
    \end{subfigure}
    \begin{subfigure}[t]{.49\linewidth}
    \includegraphics[width=1.\textwidth]{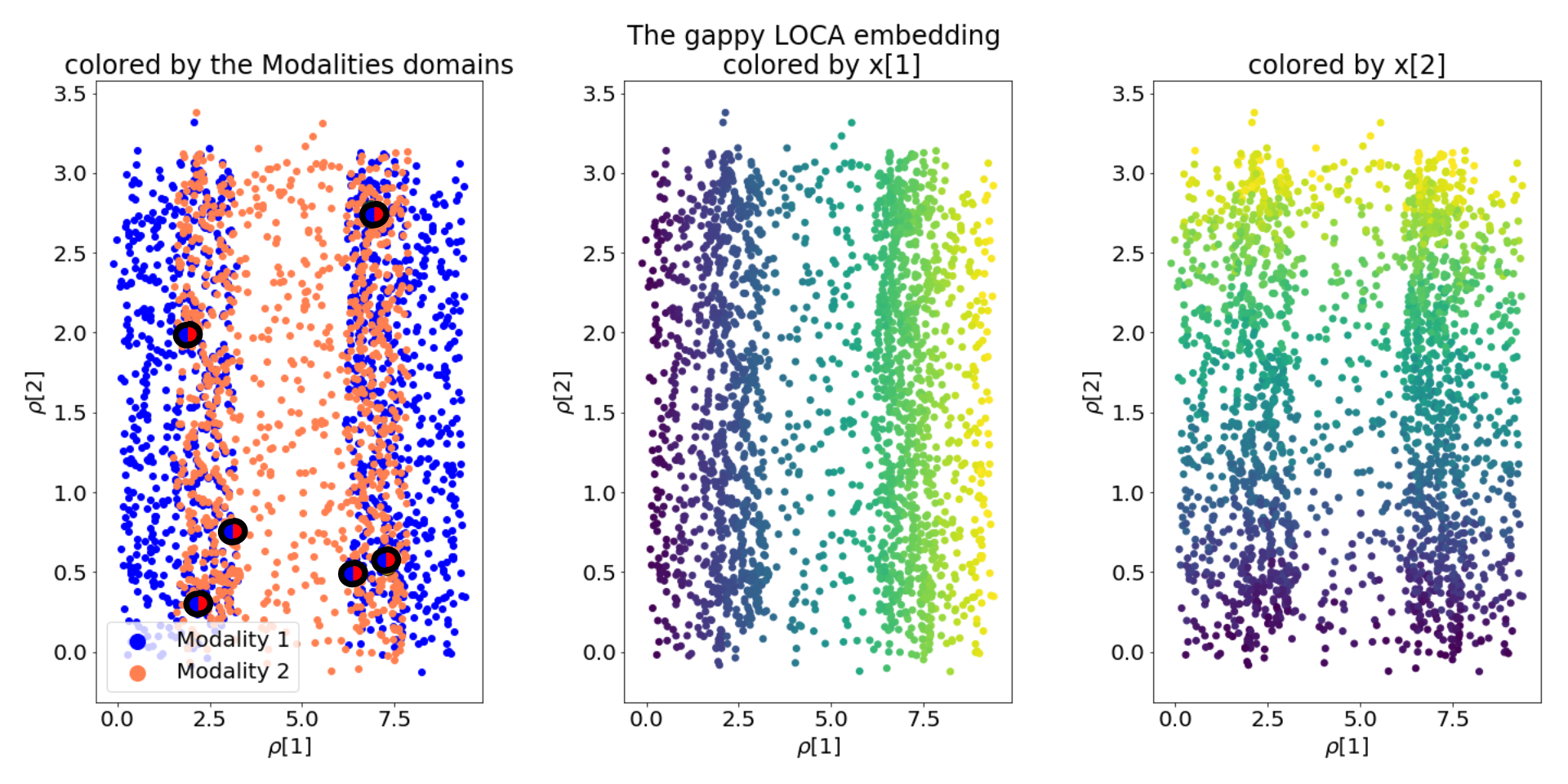}
    \caption{Gappy LOCA embedding}
    \end{subfigure}
    \caption{(a) The ``two LOCA'' embeddings, where the rigid transformation between the two was estimated based on the intersection of only one blue region and the red region. (b) The successful Gappy LOCA embedding. The cyan points are the registration burst centers for modality 1 (colored in blue), and the orange points are the registration burst centers for modality 2 (colored in red). As Gappy LOCA takes registration bursts into account during the embedding phase, it can consistently join the two disconnected regions by the orange region.}
\label{experiment_path_domain_embeddings}
\end{figure}

In Figure \ref{fig:experiments_patchy_domain} we present the observations in the observed as well as in the latent spaces. As shown on the right sub-figure, only Gappy LOCA is capable of leading to an isometric embedding in this example. A-dmaps fails in this example due to the practically disconnected, two-block structure of the incomplete affinity matrix. On the other hand, as we can see in figure \ref{experiment_path_domain_embeddings}, the LOCA embedding of the first modality produces two squares. Each of these two \textit{disconnected} squares are individually almost isometric to the original domain components; yet as the components are too far apart, the model fails to find the right relative positioning between the two. 
A qualitatively similar example, involving several (three) patches and several (two) modalities, what we call the ``French flag'' example, can be found in Appendix~\ref{appendix:french flag}.

\subsection{Experiment 4: Matrix completion}

\begin{figure}[htb]
    \centering
    \begin{subfigure}[t]{.3\textwidth}
    \centering
    \includegraphics[width=1\textwidth]{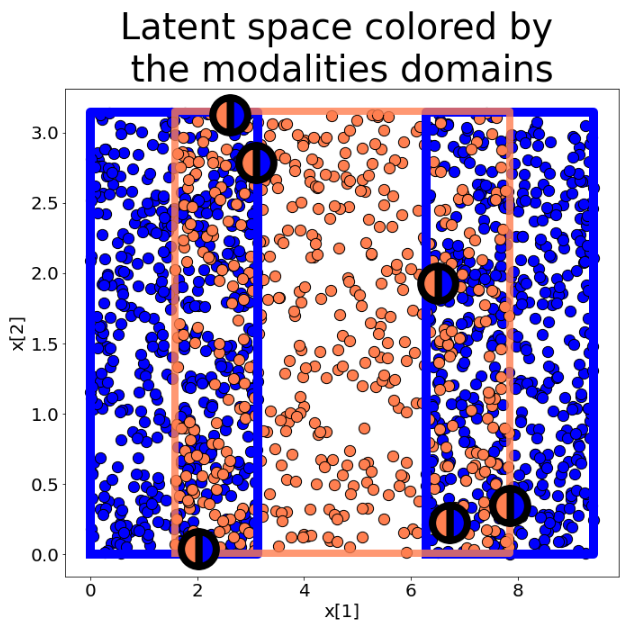}
    \caption{}
    \end{subfigure}
    \begin{subfigure}[t]{.3\textwidth}
    \centering
    \raisebox{.25cm}{\includegraphics[width=1\textwidth]{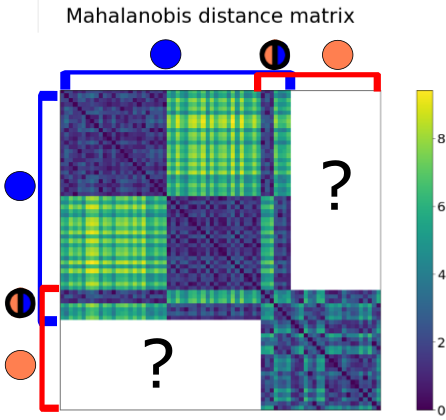}}
    \caption{}
    \end{subfigure}
    
    \caption{(a) A latent space observed by two modalities (blue and red), where the black points are the registration bursts centers that are observed by both.  (b) The extracted distance matrix. As can be seen, the black calibration points connect the red pairwise distance with the blue. The task is to uncover the missing distances between the red and blue points.
    }
    \label{fig:example_missing_dists}
\end{figure}

We again work in the setting of experiment 3, with one connected and one disconnected set of observations (see Fig.~\ref{fig:example_missing_dists}, left).
Now, however, we reinterpret the fusion of the two modality observations as a matrix completion problem. As we can only measure distances for points in each modality separately (plus a few registration points), there are large portions of the distance matrix that are simply missing (see Fig.~\ref{fig:example_missing_dists}, right).
Gappy LOCA provides a consistent and isometric embedding of the entire domain $\mathcal{X}_1\cup\mathcal{X}_2$, which we can employ to now complete (fill in) the entire distance matrix accurately.
Without completing the missing elements, the available matrix entries reporting distances between far away points (e.g., the green/yellow entries in Fig.~\ref{fig:example_missing_dists}, right) do not suffice to provide a consistent isometric embedding of the disconnected components of the domain of Modality 1 (see Fig.~\ref{experiment_path_domain_embeddings}, (a)).
%

\subsection{Experiment 5: Wi-Fi localization}

In the experiment we examine the problem of localization throughout a floor plan of an MIT building, based on the captured signals from triplets of Wi-Fi transmitters. We refer to the locations on the floor plan as our latent domain.  Each triplet (i.e., each modality) is accurately observed over a specific region along the floor, in which all its three transmitters can be captured in sufficient strength  (simulating a high signal-to-noise ratio region for each transmitter). Based only on the ensemble of received partial Wi-Fi signals, we attempt to find a consistent isometric embedding of the entire latent space (i.e., recover the entire floor plan).
\begin{figure}[H]
    \centering
    \begin{subfigure}[t]{.3\textwidth}
    \centering
    \includegraphics[trim={0 0 12cm 0},clip,height=5cm]{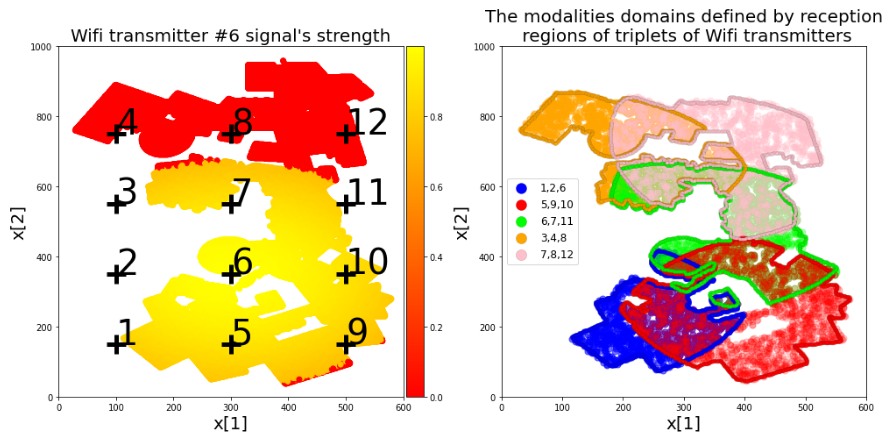}
    \caption{}
    \end{subfigure}
    \begin{subfigure}[t]{.3\textwidth}
    \centering
    \includegraphics[trim={12cm 0 0 0},clip,height=5cm]{figs/Wifi/config.PNG}
    \caption{}
    \end{subfigure}
    \begin{subfigure}[t]{.3\textwidth}
    \centering
    \includegraphics[trim={8cm 0 0 0},clip,height=5cm]{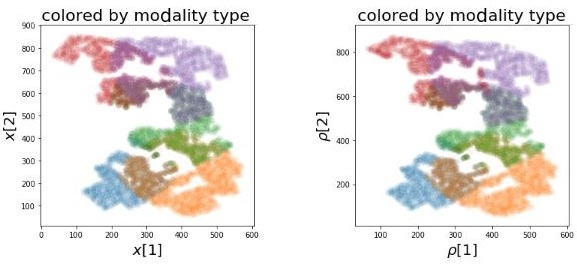}
    \caption{}
    \end{subfigure}

    \caption{(a) The latent domain with 12 Wi-Fi transmitters overlayed, the color indicates the signal strength received in each location from Wi-Fi transmitter-$6$. 
    (b) The latent domain separated into the different modalities' domains, each defined upon three specific Wi-Fi transmitters (indicated in the legend). Each modality's domain contains locations in which the signal's strength from all three Wi-Fi transmitters is strong enough. (c) Gappy LOCA embedding, up to a global rigid transformation.}
    \label{fig:wifi}
\end{figure}

We use $12$ Wi-Fi transmitters placed at different positions on the floor plan (see Figure \ref{fig:wifi}). The strength of each Wi-Fi received signal is modeled by
\begin{eqnarray*}
s_i(\V{x})= 
\exp\left( - \frac{\| \V{x} -\V{\mu}_i\|_2^2 }{100^2}\right)   \qquad \qquad \qquad i\in \{0,\ldots,11\},
\end{eqnarray*}
where $\V{\mu}_i$ is the $i$-th transmitter's location and $\V{x}$ is a location on the floor plan where $x\in\mathbb{R}^2$. In order to make the experiment more realistic, we further threshold weak signals based on
\begin{eqnarray*}
g_i(\V{x})= \left\{ \begin{array}{cc}
s_i(\V{x}) & s_i(\V{x})\geq \text{Threshold} \\
0 & else
\end{array}\right.  \qquad \qquad i\in \{0,\ldots,11\}.
\end{eqnarray*}
We motivate this thresholding operation as a setting in which the Signal-to-Noise ratio is too low to usefully exploit the signal.
Modalities (triplets) are only usable when all three receivers are characterized by a high received signal; The physical points where this occurs define each modality's domain. Specific triplets of Wi-Fi-transmitters that we examine are $(1,2,6)$, $(5,9,10)$, $(6,7,11)$, $(3,4,8)$ and $(7,8,12)$. 
In the right panel of Fig.~\ref{fig:wifi} we distinctively color the different domains of the different modalities, i.e., locations in which all three of the modality's Wi-Fi transmitters are presumed to be strong enough. These modalities are somewhat arbitrarily prescribed. So are the areas in which each of their constituent transmitters is assumed to be above threshold, see, e.g., Fig~\ref{fig:wifi}(a) for ``strong reception area'' for transmitter 6.
The $5$ modalities are modeled as non linear functions of points in the latent domain  by:
\begin{gather*}
\V{f}^{1} (\V{x}) = \left(\begin{array}{c} 
g_1(\V{x}) \\ g_2(\V{x}) \\ g_6(\V{x})
\end{array} \right) \qquad 
\V{f}^{2} (\V{x}) = \left(\begin{array}{c} 
g_5(\V{x}) \\ g_9(\V{x}) \\ g_{10}(\V{x})
\end{array} \right) \\
\V{f}^{3} (\V{x}) = \left(\begin{array}{c} 
g_6(\V{x}) \\ g_7(\V{x}) \\ g_{11}(\V{x})
\end{array} \right) \qquad 
\V{f}^{4} (\V{x}) = \left(\begin{array}{c} 
g_3(\V{x}) \\ g_4(\V{x}) \\ g_8(\V{x})
\end{array} \right) \qquad
\V{f}^{5} (\V{x}) = \left(\begin{array}{c} 
g_7(\V{x}) \\ g_8(\V{x}) \\ g_{12}(\V{x})
\end{array} \right). \qquad 
\end{gather*}
%

We apply Gappy LOCA to this data using registration point bursts located at the intersections between different modalities. The embedding that we obtain is presented in Figure 2 along the original latent domain.

\subsection{Experiment 6: Wave equation solution with domains in the form of Maryland's districts}
\begin{figure}[h!]
    \centering
    \includegraphics[width=1\textwidth]{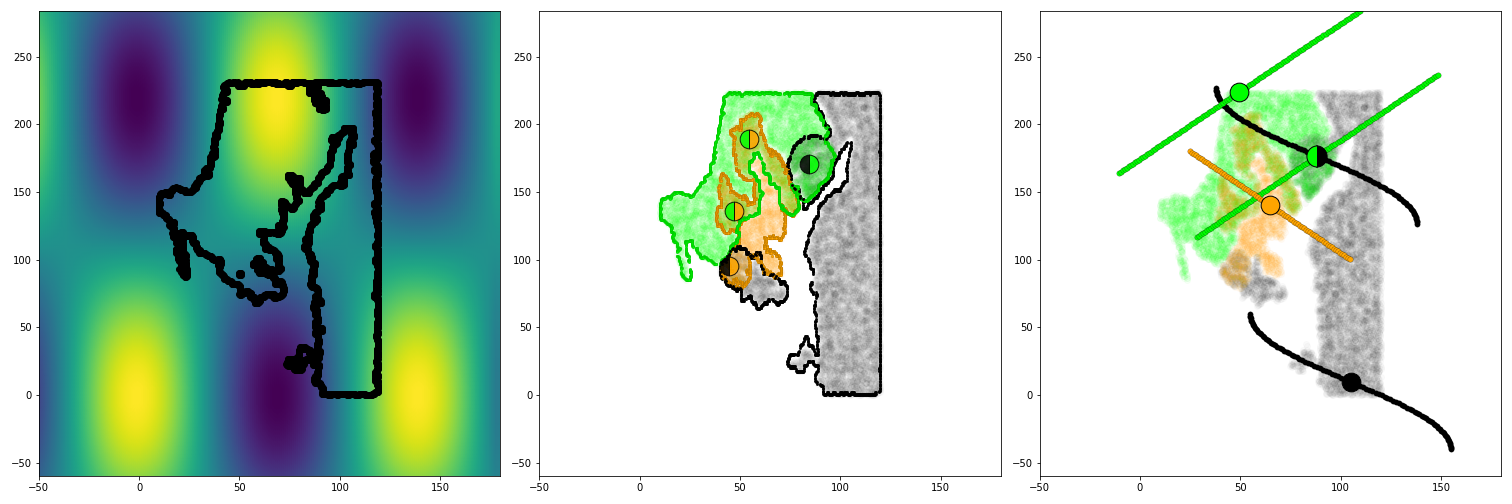}
    \includegraphics[width=1\textwidth]{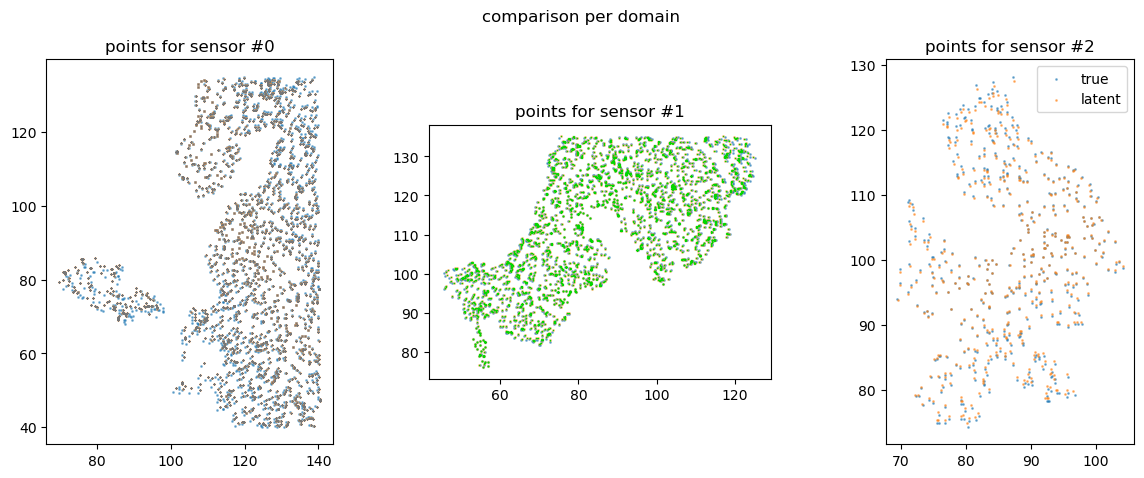}
    \caption{Top row, left: The latent space colored based on \eqref{eq:maryland}, the faded region includes points that were not sampled. Center: The latent space domain colored by the modalities domain. Right: Three types of modalities are being used to sample the latent space in the form of trajectories on the latent space that crosses the sampled point exactly at their center. These trajectories include the value of the equation throughout the trajectory, meaning that they do not have locations in the latent space. Bottom row: the reconstructed domains, compared to the actual points. Note that Gappy LOCA constructs all of these simultaneously, we just show them separately for easier comparison.}
    \label{fig:KS_configuration}
\end{figure}


In this final experiment, we use Gappy LOCA to piece together data collected from recording the solution of a wave equation by observing three different parts of the spatial domain, and through different observations in each of the three parts. The wave equation is a widely known partial differential equation arising in the modeling of many different physical phenomena, from acoustics to fluid flows. The specific solution that we consider is
\begin{eqnarray}
u(x,t) = \cos\left(\frac{2\pi x}{140}\right) \cos\left( \frac{2\pi t}{440}\right),
\label{eq:maryland}
\end{eqnarray}
where $x\in [0,2\pi)$ denotes the one-dimensional space variable  and $t\in [1,1.4]$ denotes the time variable. 
In our example, the data are collected by several different travelling observers; 
they are gathered without knowledge of the specific locations in space or points in time at which the short trajectories were captured. We only know the measurements produced by each of the three differently travelling observers.
%
%
The task is to uncover the global space-time geometry from this ensemble of short observed trajectories.

Formally, we define the domain of the PDE solution by $\mathcal{X}\subset \mathbb{R}\times \mathbb{R}$ using the space and time values, and the PDE solution by $u:\mathcal{X}\rightarrow \mathbb{R}$, illustrated on left plot of \figref{fig:KS_configuration}. 
Next we define three types of measurement devices (travelling observers) that sample the solution $u$. Their irregular domains are characterized by the mid point in space/time of each trajectory; the (irregular and fragmented) domains of the three types of travellers are denoted by $\mathcal{X}_1,\ldots,\mathcal{X}_3$ and illustrated in the second plot in \figref{fig:KS_configuration}. Note that these irregular and fragmented domains possess intersections in which a few registration points are jointly observed by more than one travelling observer type. The specific modalities are given in the Appendix, and are illustrated in the right of \figref{fig:KS_configuration}. 

\section{Discussion}\label{sec:discussion}

The essential feature in this paper that allows for successful fusion of different sensor measurements---different partial observations of the same object (the same experiment, the same process)---is the assumed availability of not just single observation for each measured instance, but rather of a measurement ``burst" around the instance: a set of slightly perturbed nearby instances. This set of nearby instances has to be rich enough to allow the estimation of the {\em local} Jacobian of the observation/measurement function of the object.
Knowledge of these local Jacobians allows us to invert the measurement function and obtain a ``Platonic" representation of the object.
We showed how to extend this principle from inverting a single measurement function (as in~\cite{singer2008non,peterfreund2020local}) to the consistent inversion of multiple partial measurement functions towards a {\em common, global} Platonic representation. 
In effect, after individual observation functions are inverted, the remaining challenge is to {\em correctly rigidify} all of them in
a consistent latent space.
Doing this is equivalent to completing the pairwise Mahalanobis distance matrix across all pairs of points, independent of the
available partial observation set(s) for each point. 
The approach was implemented in a single, end-to-end pipeline
in the form of a multiple-auto-encoder neural network architecture;
the inputs to the pipeline are the sets of partial observations of each point, and the result is a globally consistent latent space harmonizing (rigidifying, fusing) all measurements.

The biggest computational challenge/obstacle to training this architecture proved to be the difficulty in consistently initializing
the auto-encoder ensemble.
The method ensures that individual encoder representations are invariant to diffeomorphisms up to orthogonal transformations: rotations, translations and reflections. 
Rotations and translations are continuous groups, and appear to be ``easily learnable" through stochastic gradient descent.
The problem proved to be the (discrete) reflections: different ``patches" of data, during initial stages of training, ``select" reflections that are not aligned, and this causes the optimization process to get trapped.
While we initially considered partially training the encoders and then
solving a linear alignment problem (\cite{singer-2012}), what proved to work more consistently for us was to embed the data in higher dimension than strictly necessary, so that reflections can be represented by (higher dimensional) rotations, removing their discrete nature and thus facilitating optimization/training using iterative methods (here, SGD).
We demonstrated the approach in examples of increasing complexity, including a Wi-Fi localization example, as well as ``dynamical puzzle" example: the fusion of different types of observation of the solutions of a Partial Differential Equation. In the latter context, what we do can be considered as a form of data assimilation for dynamical systems across multiple data sources.

In our view, the key insight is that multiple measurements of the 
same instance are crucial in facilitating the construction of a global consistent representation. Typical processing of multiple measurements (e.g. involving averaging, so as to reduce variance) may be less helpful than using local measurement ensembles as a gauge of the distortion induced by the observation function -- our main enabling tool. It appears to us that instruments designed to collect, and even induce, multiple measuerements corresponding to our ``bursts", would make the computational data fusion task much more systematic and efficient. Given such instruments, the problem of data fusion is reduced to rigidification of the measurements -a linear problem in the right latent space- and (remembering Erasmus) the process could be termed {\em Rigiditae Encomium} - ``In Praise of Rigidity".

\begin{acknowledgements}
  The work of YGK, RRC, EP and OL was partially supported by the Defense Advance Research Projects Agency; FD and IB were partially funded by the Deutsche Forschungsgemeinschaft (DFG, German Research Foundation) – project no. 468830823. 
\end{acknowledgements}

\bibliography{paper}
\bibliographystyle{plain}

\appendix

\section{French flag example}\label{appendix:french flag}
\begin{figure}[ht!]
    \centering
    \includegraphics[width=1.\textwidth]{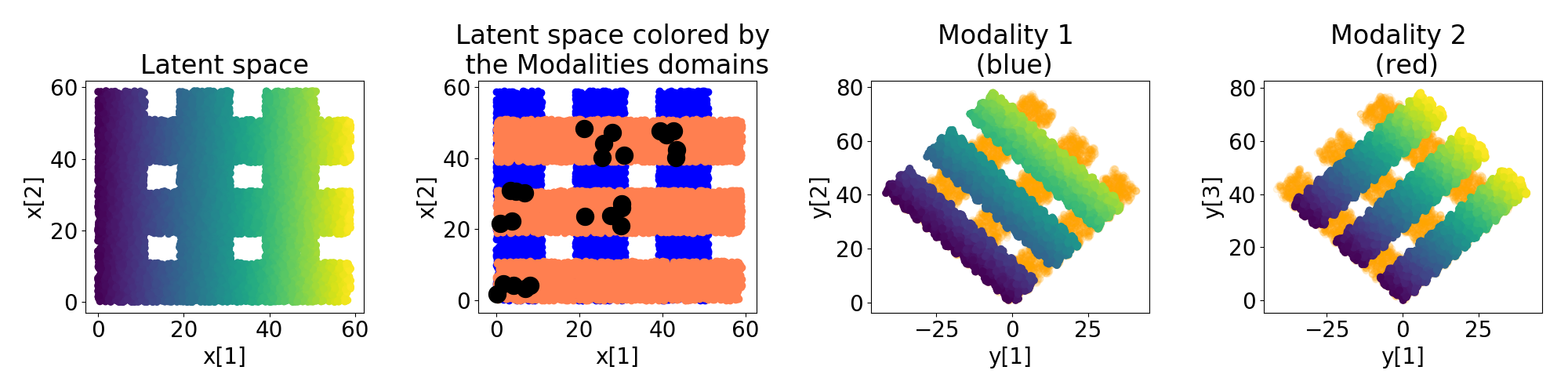}
    \caption{Multiple observers with patchy domain}
    \label{fig:woven mat french flag}
\end{figure}

Fig.~\ref{fig:woven mat french flag} illustrates that several internally disconnected patches can also be combined into a coherent embedding. 

\section{Experiment 2.5 - a point case}\label{appendix:point case}

%
%
%

\begin{figure}[ht!]
    \centering
    \begin{subfigure}[b]{0.25\textwidth}
        \centering
        \includegraphics[width=1\textwidth]{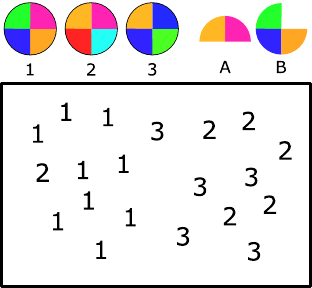}
        \caption{}
    \end{subfigure}
    \begin{subfigure}[b]{0.25\textwidth}
        \centering
        \includegraphics[width=1\textwidth]{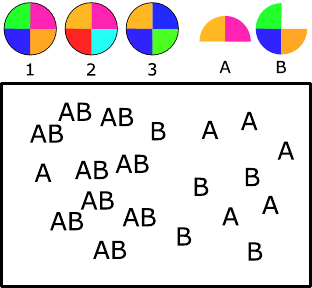}
        \caption{}
    \end{subfigure}
    \caption{An illustration of the ``point case''. (a) There are three different types of points (1, 2, 3), which are measured by four sensors each (colors). It is not possible to define each point type as a modality, as, for example, type 2 has an instance on the very left of the domain, far away from the other points of the same type. (b) However, it is possible to define new modalities by subsets of sensors (A,B). Then, the Gappy LOCA approach can be applied to these joint modalities. For point type 1, both modalities are present for each instance.}
    \label{fig:flowers-illustration}
\end{figure}

Up to now, we considered a problem in which we already knew how many modalities could be used. We also had a reasonable sense of the patches that each of these modalities could rigidify. In the more general case, this information must be deduced from knowing how many independent sensors are observing each point (so, in some sense, every point is a patch, and is common across all modalities that are observing it). 
Our search algorithm will then tell us whether every ``point-patch'' is rigidifiable with enough other ``point-patches'' and which combination of sensors will accomplish this for every single point.
In a sense the ``point'' version of Gappy LOCA reduces to the ``patch'' version by considering each point as a patch. The theoretical count of possible modalities is  ``N factor k'', where N is the total number of sensors, and k is the sufficient ``Whitney'' minimal number of sensors to connect a point to another point.
%
Conceptually, one would have a factorial number of encoders based on this reasoning, and each ``point-patch'' would be an input to all rigidifying modalities it participates in.
One envisions instead an additional preprocessing step in Gappy LOCA that selects--when rigidification is possible--a minimal number of encoders (rigidifying modalities).
Now every point is a ``common point'' for the rigidifying modalities it participates in. 
Fig.~\ref{fig:flowers-illustration} illustrates this in an example.
The registration loss should now be computed for all points in latent space that are arrived at through more than one encoder.

\end{document}